\documentclass{article}

\usepackage[nonatbib,final]{nips_2018}

\usepackage[utf8]{inputenc} % allow utf-8 input
\usepackage[T1]{fontenc}    % use 8-bit T1 fonts
\usepackage{hyperref}       % hyperlinks
\usepackage{url}            % simple URL typesetting
\usepackage{booktabs}       % professional-quality tables
\usepackage{amsfonts}       % blackboard math symbols
\usepackage{nicefrac}       % compact symbols for 1/2, etc.
\usepackage{microtype}      % microtypography

\usepackage{graphicx}
\usepackage{tabularx}
\usepackage{amsmath}
\usepackage{amsthm}
\usepackage{amssymb}
\usepackage{hyperref}

  \usepackage[caption=false,font=footnotesize]{subfig}

\usepackage{booktabs}
\usepackage{multirow}
\usepackage{graphicx}
\usepackage[export]{adjustbox}
\usepackage[table,xcdraw]{xcolor}
\usepackage[normalem]{ulem}
\useunder{\uline}{\ul}{}

\usepackage{algorithm}
\usepackage[noend]{algpseudocode}
\makeatletter
\def\BState{\State\hskip-\ALG@thistlm}
\makeatother

\usepackage[font=footnotesize,labelfont=bf]{caption}

\theoremstyle{plain}

\theoremstyle{definition}

\graphicspath{ {images/} }

\title{What is Interpretable? Using Machine Learning to Design Interpretable Decision-Support Systems}

\author{Owen~Lahav\\
  Department of Computer Science\\
  University of Oxford\\
  OX1 3QD UK\\
  \texttt{oren.lahav@gtc.ox.ac.uk}
  \And
  Nicholas Mastronarde \\
  Department of Electrical Engineering \\
  University at Buffalo \\
  Buffalo, 14228, NY, USA \\
  \texttt{nmastron@buffalo.edu} \\
  \And
  Mihaela van der Schaar \\
  Department of Engineering Science \\
  University of Oxford\\
  OX1 3QD UK\\
 \texttt{mihaela.vanderschaar@oxford-man.ox.ac.uk}
}

\begin{document}

\maketitle

\begin{abstract}
Recent efforts in Machine Learning (ML) interpretability have focused on creating methods for explaining black-box ML models. However, these methods rely on the assumption that simple approximations, such as linear models or decision-trees, are inherently human-interpretable, which has not been empirically tested. Additionally, past efforts have 
focused exclusively on comprehension, neglecting to explore the trust component necessary to convince non-technical experts, such as clinicians, to utilize ML models in practice.
%neglected to model the user-designer interactions necessary to convince non-technical experts, such as clinicians, to utilize ML models in practice.
In this paper, we posit that reinforcement learning (RL) can be used to \emph{learn} what is interpretable to different users and, consequently, build their trust in ML models. 
To validate this idea, we first train a neural network to provide risk assessments for heart failure patients. We then design a RL-based clinical decision-support system (DSS) around the neural network model, which can learn from its interactions with users. We conduct an experiment involving a diverse set of clinicians from multiple institutions in three different countries. Our results demonstrate that ML experts cannot accurately predict which system outputs will maximize clinicians' confidence in the underlying neural network model, and suggest additional findings that have broad implications to the future of research into ML interpretability and the use of ML in medicine.

%Our results lead to interesting new findings on interpretability and have broad implications across ML-driven systems, from DSSs for medicine to autonomous vehicles.
%how to best present the underlying neural network model and resulting risk assessments to the clinicians in order to maximize their confidence in the system.

%We posit that reinforcement learning (RL) techniques can be utilized to achieve interpretability empirically. By constructing a RL-based survey tool and conducting an experiment with medical and ML experts we are able to design a demonstrably interpretable, useful, and usable decision-support system based on a neural network that provides risk assessments for heart failure patients. We believe our study lays the foundations for future work in interpretability that will enable the ML research community to create models and systems that are demonstrably effective and have high user trust.
\end{abstract}

\section{Introduction}

\textbf{Motivation:} 
Machine Learning (ML) models have been shown in many cases to achieve higher predictive power than simpler statistical methods \cite{waljee_higgins_2010}. It is therefore easy to envision sophisticated, carefully calibrated ML models being utilized to support critical decision-making as in, e.g., \cite{esteva_2017}, \cite{gulshan_2016}. However, despite their ability to provide accurate predictions, ML models have not been heavily utilized in fields such as medicine and prognostic research \cite{cleophas_zwinderman_2015} \cite{kononenko_2001} \cite{cabitza_rasoini_gensini_2017}. 

One reason for this is the inherent complexity of black-box ML models like neural networks. Such models are powerful because of their ability to detect complex patterns in data, achieving high predictive accuracy. However, this makes them difficult to explain to non-experts in less technical fields. Users wishing to leverage predictive models for critical decision-making, such as medical risk prognosis, must be professionally and ethically able to justify their medical actions, explicitly linking inputs like patient characteristics to predicted outcomes \cite{obermeyer_emanuel_2016}. Existing neural networks and other ML models do not readily provide for this. As a result, the ML research community has been paying increased attention to interpretability.

We define ML model interpretability as the extent to which a ML model can be made understandable to relevant human users, with the goal of increasing users' trust in, and willingness to utilize, the model in practice. We argue that this focus on trustworthiness has been neglected from past literature but is absolutely vital - a comprehensible model is not useful unless it is also trusted. A more extensive discussion can be found in Appendix \ref{AppInt}.

A recent survey compiled an overview of methods used to explain black-box ML models \cite{DBLP:journals/corr/abs-1802-01933}. To date, most research on ML interpretability focuses on developing methods we term {\bf interpretability modules}, which run alongside existing black-box models to produce statistical explanations that are generally accepted as being easily understandable to humans. These methods range from ranking input features' contributions to generating entirely new models that closely approximate the original black-box while using a simpler, supposedly easier-to-understand methodology such as linear regression (e.g., LIME \cite{ribeiro_singh_guestrin_2016}, SLIM \cite{slim}), decision-trees \cite{Craven:1995:ETR:2998828.2998832} or logic rules \cite{luo_2016}.

These studies rely on important assumptions. First, it is assumed that feature rankings, linear models, or decision-trees are \emph{indeed} interpretable, and that they consistently increase the utility of ML models for clinical users. This assumption appears overly simplistic -- it is likely that interpretability is \emph{subjective}, requiring different approaches for different users and in different contexts. Secondly, it is assumed that simply presenting the modules is enough to achieve interpretability. We argue that this is \emph{not} enough: instead, interpretable systems require a \emph{user-in-the-loop} design approach based on an \emph{interactive} process enabling designers to convince people to trust and utilize ML-driven systems.
% -- ranging from decision-support systems to autonomous vehicles.
%the users should be involved in the development of interpretable systems relying on ML models, and it is through an \emph{interactive} process that designers can convince people to trust and make use of ML-driven systems -- ranging from decision-support systems to autonomous vehicles.

Limited work has investigated these assumptions and how different types of ML model evidence actually affect users' trust \cite{Yin2018} \cite{Collaris2018}.
Our work differs from these recent efforts as we propose to use Reinforcement Learning (RL) to present different methods of interpretability and thus \emph{learn} what is effectively interpretable to different users and, consequently, build their trust in ML models.

\textbf{Contributions:}
We first train a neural network to provide risk assessments for heart failure patients. We then design a RL-based clinical decision-support system (DSS) around the neural network model. The DSS presents a sequence of interpretability modules and other forms of evidence about the underlying ML model to the user (e.g., information about the data-set, training methodology, and model accuracy, in addition to interpretability modules including local linear and decision-tree approximations). As users interact with the DSS, it learns to present an information sequence that maximizes users' expected trust in the ML model using RL. 
We asked 14 clinicians\footnote{The authors would like to thank the medical doctors who have provided crucial consultations and participated in our experimental study. A partial list including affiliations can be found in Appendix \ref{contributers}.} and 30 ML experts from multiple institutions and countries to interact with our DSS. Clinicians rated their trust in the ML model as they used the system, while ML experts indicated if they believe that the presented information would increase the average clinician's trust in the ML model. Our results lead to new findings that may have broad implications for the future of ML interpretability.

\section{Interpretable Decision-Support System Design}
\label{DSS}

\textbf{Data-set and Model:} 
Given the prevalence of heart disease globally, we elected to focus on developing a DSS to predict 1-year mortality risk for heart failure patients. We utilize the Meta-analysis Global Group in Chronic Heart Failure data-set, or \textbf{MAGGIC}, first used in the meta-analysis study of Popock et al. \cite{pocock_etal._2012}. Leveraging patient data from 30 cohort studies, the data-set includes 30,389 heart failure patients, of whom 18.8\% died within 1 year. 
The data-set contains 31 features per patient including patient characteristics such as age and body mass index (BMI), physical symptoms such as shortness of breath, and prescribed medications including ACE Inhibitors and Beta-blockers. Missing values were imputed using Multivariate Imputation by Chained Equations (MICE) \cite{buuren_groothuis-oudshoorn_2011}.

Using this data-set, we trained several ML models, summarized in Table \ref{table:Preds}. We selected a simple implementation of a deep \textbf{neural network}, comprising two fully-connected layers of 100 and 20 nodes, which achieves good predictive performance among the other tested methods, including the MAGGIC Risk Score \cite{pocock_etal._2012} that is currently used by clinicians. Since deep learning is increasingly popular and gaining interest in the medical community, yet is clearly a black-box without clearly comprehensible parameters, it is a good fit for our study.
% exploring the factors that influence clinicians' confidence and willingness to use sophisticated yet opaque ML models.

%Our particular implementation of the neural network was naive, consisting of two layers of 100 and 20 nodes respectively, ReLU activation, and no dropout, achieving an area under the receiver operating curve (AUC-ROC) of 0.725. We are well-aware that novel, exotic architectures could have led to more accurate models; however, our aim was to demonstrate the utility of using RL to learn interpretability of \emph{any} black-box model, not only the most carefully calibrated, accurate networks. 

\begin{table*}[h]
\caption{Predictive performance for various models on the MAGGIC data-set \protect\cite{pocock_etal._2012}}
\centering
\resizebox{0.55\columnwidth}{!}{
\begin{tabular}{lll}
\rowcolor[HTML]{C0C0C0} 
{\ul \textbf{Model}} & {\ul \textbf{AUC-ROC}} & {\ul \textbf{AUC-PR}} \\
\textbf{Linear Regression} &  $0.573\pm0.0078$ &  $0.250\pm0.0023$ \\ \hline
\textbf{Random Forest} & $0.731\pm0.0046$ & $0.328\pm0.0105$  \\
\textbf{Gradient Boosting Machine} &  $0.710\pm0.0031$ & $0.373\pm0.0116$ \\
\textbf{XGBoost} &  $0.711\pm0.0041$ & $0.371\pm0.1110$ \\
\textbf{Neural Network} &  $0.725\pm0.0054$ &  $0.376\pm0.0060$\\ \hline
\textbf{MAGGIC Risk Score} &  $0.693\pm0.0071$ & $0.324\pm0.0121$ 
\end{tabular}
}
\label{table:Preds}
\end{table*}

\textbf{Model Evidence and Expert Consultations:}
%Having developed a ML model in the form of a neural network, we proceeded to collect model evidence to present to users as part of a DSS. 
We aimed to identify the sequence of evidence that maximizes expected clinician trust in our DSS and the underlying ML model. 
%As shown in Table \ref{table:ModelEvidence}, 
Evidence can include general model information such as accuracy and architecture, individual predictions, and interpretability modules. 
%Based on this list, we identified different types of evidence we could present pertaining to our neural network model.
Since the set of possible evidence is quite large, we applied \emph{expert heuristics} to reduce the size of our search space. We consulted with \textbf{3 doctors}, presenting them with lists of possible model evidence and asking for professional opinions regarding their relevance. We used these expert heuristics to eliminate the evidence that was judged to be unlikely to increase clinicians' trust in the ML model. Though critical for reducing our search space to a tractable size, expert consultations may have introduced unintentional bias. Future experiments may elect to consult additional clinicians or forgo the use of heuristics to reduce bias.
Table \ref{table:InfoSummary} summarizes the set of model evidence we included in our DSS. These have been divided in two parts to further reduce sample complexity: Overall Model Evidence and Specific Prediction Evidence. More details can be found in Appendix \ref{AppA}.

\begin{table*}[h]
\caption{A summary of the model information presented as part of our DSS.}
\centering
\resizebox{0.95\textwidth}{!}{
\begin{tabular}{@{}lll@{}}
\toprule
\textbf{Survey Part} & \textbf{Model Information} & \textbf{Details} \\ \midrule
\multirow{5}{*}{\textbf{\begin{tabular}[c]{@{}l@{}}Part 1: \\ Overall Model\\ Information\end{tabular}}} & Data & Data-set size, features list, feature statistics \\ \cmidrule(l){2-3} 
 & Model Methodoloy & Training and implementation: Cross-validation, neural network architecture \\ \cmidrule(l){2-3} 
 & Model Accuracy & \% Accuracy, AUC-ROC, PR-ROC \\ \cmidrule(l){2-3} 
 & Stratified Linear Approximations & Significant features of linear approximation models for each risk quintile (0-20\%, 20-40\%, etc.) \\ \cmidrule(l){2-3} 
 & Decision Tree Approximations & Diagram of decision tree approximation models for each risk quintile (0-20\%, 20-40\%, etc.) \\ \midrule
 
\multirow{5}{*}{\textbf{\begin{tabular}[c]{@{}l@{}}Part 2:\\ Specific Prediction \\ Information\\ (For Data points/\\ patients in test set)\end{tabular}}} & Prediction & Individual patient characteristics and resulting risk score for an individual data point/patient \\ \cmidrule(l){2-3} 
 & Local Feature Sensitivity (Interactive) & Summary of how different patient features affect predicted risk score - interactive DSS component \\ \cmidrule(l){2-3} 
 & Patient Outcome & Did patient survive 1 year after heart failure event? (Was prediction accurate?) \\ \cmidrule(l){2-3} 
 & Local Linear Approximation & Coefficients of a local linear approximation model \\ \cmidrule(l){2-3} 
 & Local Decision Tree Approximation & Diagram of local decision tree model \\ \bottomrule
\end{tabular}
}
\label{table:InfoSummary}
\end{table*}

\textbf{Reinforcement Learning Approach:}
In order to identify the sequence of model evidence that results in maximal expected user confidence, we model the problem as a multi-armed bandit \cite{best-arm-identification-multi-armed-bandits} and solve it using the well-known Upper Confidence Bound (UCB1) algorithm \cite{auer2002}, which is easy to implement in an online survey environment and has better sample complexity than naive approaches.
%, which efficiently balances exploration and exploitation. 
In this context, we view relevant subsequences of evidence from Table \ref{table:InfoSummary} as arms, and a user's rating of their trust/confidence in the model on a scale of 1-5 as the reward.
%we view each sequence as an arm, and expected confidence as a reward. Our problem becomes a best-arm identification within the Multi-Armed Bandit paradigm \cite{best-arm-identification-multi-armed-bandits}. A widely accepted methodology for solving the best-arm identification problem is the Upper Confidence Bound (UCB1) algorithm \cite{auer2002}. The algorithm's success stems from the fact that it balances exploration and exploitation efficiently.
We note that other algorithms for solving such RL problems exist, and can be utilized for addressing the DSS design problem. Future work may do well to experiment with other methods. 
%as part of our overall approach of applying RL to interpretability.

%In order to present different sequences of evidence to doctors and capturing their resulting confidence in the system and underlying model, we have constructed the Multi-Armed Survey Tool Enabling Reinforcement (MASTER), an online tool implementing UCB1. The survey tool is capable of:
%\begin{itemize}
%\item{Tracking multiple arms and their respective estimated mean confidence and upper bound}
%\item{Displaying model evidence, individual predictions, interpretability module results, and other information based on selected arms}
%\item{Prompting clinicians to report their confidence in the system}
%\item{Updating the expected confidence and upper confidence bound per arm accordingly}
%\end{itemize}
%Fig. \ref{fig:MASTER} represents a block diagram of the tool and methodology.
%
%\begin{figure}[!h]
%\centering
%\includegraphics[width=0.5\columnwidth]{MASTER}
%\caption{A block diagram showcasing MASTER. Each new respondent to the survey tool triggers the UCB1 algorithm, and the arm with the current highest UCB is selected. The arm's respective model evidence is shown to the respondent, who identifies his/her resulting confidence in the model. This reward is used to update the UCB values in a feedback loop, which will influence the arm and information shown to the next respondent}
%\label{fig:MASTER}
%\end{figure}

%You may find our tool online at:
%\begin{itemize}
%\item{\textbf{Medical Version}: \url{https://mlindecisionsupport.github.io/}}
%\item{\textbf{Computer Scientist Version}: \url{https://mlinterpretability.github.io/}}
%\end{itemize}

\section{Experimental Results}
\label{exp}

%1. Intro to results
We collected responses from 14 medical doctors who used our DSS and indicated their confidence in the system and underlying neural network model. In addition, we sought to compare the doctors' results with the current beliefs of ML experts, and thus we asked 30 computer scientists with ML backgrounds to indicate their expectations regarding an average doctor's confidence score. All responses have been normalized to the range $[0, 1]$, with 1 representing maximum confidence. The DSSs presented to clinicians and ML experts are available online at \cite{master_clinician} and \cite{master_cs}, respectively.

%2. Summary of arms

%\textbf{Best Sequence Identification:}
%First, we attempted to identify the best evidence sequences (``arms'') for doctors and computer scientists, respectively. Fig. \ref{fig:ResultsArms} presents the mean user confidence and upper confidence bound for each arm for both doctors and computer scientists. We draw several insights.

\begin{figure}[!t]
\centering
\subfloat[Part 1: General Model Evidence sequences]{\includegraphics[width=0.45\columnwidth]{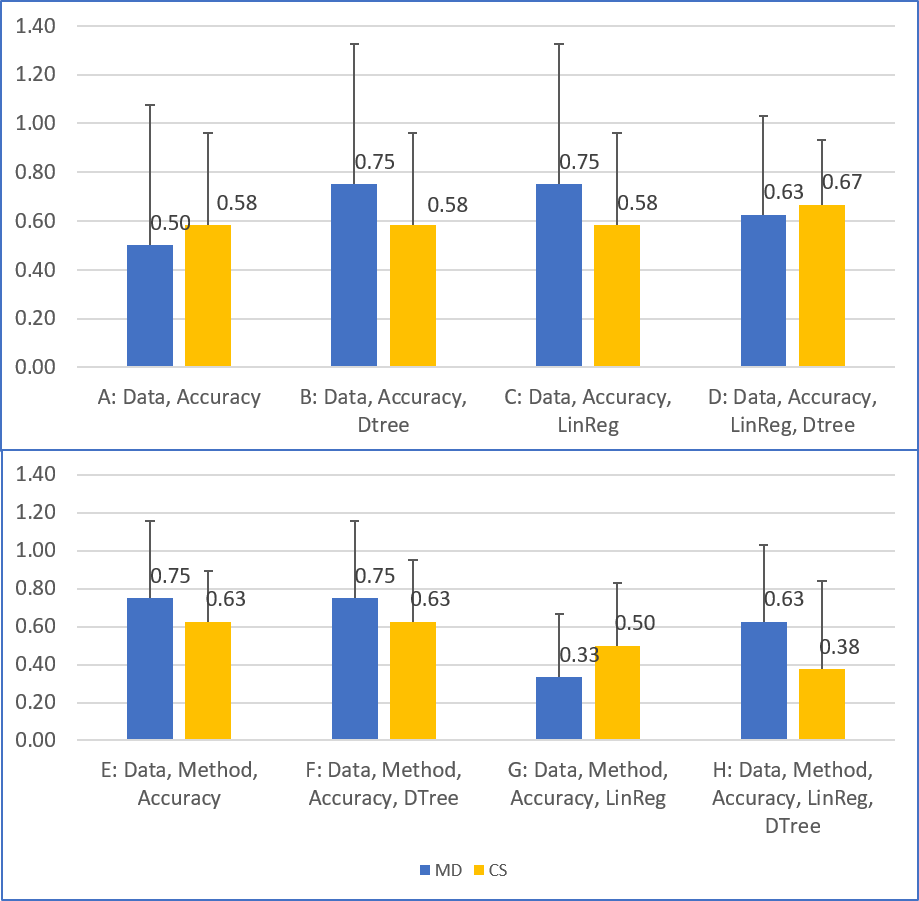}%
\label{fig:ResultsArms1}}
\hfil
\subfloat[Part 2: Patient Scenario sequences]{\includegraphics[width=0.45\columnwidth]{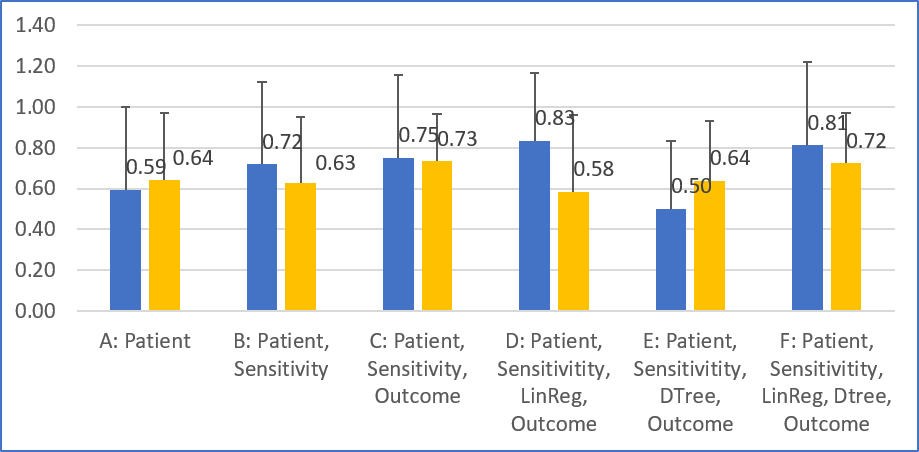}%
\label{fig:ResultsArms2}}
\caption{\label{fig:ResultsArms} Average confidence rating per arm/sequence. Blue bars denote the doctors' responses and orange bars denote the ML experts' expectations. Error bars indicate the upper confidence bounds as calculated by the UCB1 algorithm. (a) Part 1: General Model Evidence; (b) Part 2: Specific Patient Scenarios.}
\end{figure}

%We can draw an additional key insight - the results and patterns for medical doctors and computer scientists appear to be \emph{different}. 
Fig.~\ref{fig:ResultsArms} presents the mean user confidence rating and upper confidence bound for each evidence sequence (``arm'') for both doctors and computer scientists. First, Fig. \ref{fig:ResultsArms1} reveals that ML experts are unable to accurately predict how different arms will affect doctors' confidence. In particular, ML experts expect that sequence D, containing all evidence except methodology, will maximize doctors' confidence; however, doctors show higher confidence based on four other arms. This supports our core hypothesis that interpretability requires an \emph{interactive} process between users and system designers. Simply presenting an interpretability module does not suffice -- users must be consulted to verify interpretations are acceptable, leading to utilization of the model in practice.

Fig. \ref{fig:ResultsArms2} reveals similar results. Focusing on arms A-C, the ability to interact with a DSS and examine how different features change a patient's risk score, representing feature sensitivity, appears to have a high impact on doctors' confidence. On the other hand, adding patient outcomes resulted in a smaller, marginal increase in confidence. This is in sharp contrast to ML experts, who predicted the opposite trend. This further supports the hypothesis that there is a divergence between what ML experts believe will lead to interpretability and results in practice.

Second, we find that \emph{more} evidence does not always improve users' trust, despite our initial hypothesis that information is super-additive. In fact, the longest sequence of evidence in Part 1 (H) results in \emph{lower} average confidence than shorter arms. This suggests `information overload' is possible, i.e., more evidence may cause confusion, reducing trust \cite{miller1956magical}. Thus, attempts at achieving interpretability should not simply provide \emph{more} information, but rather \emph{more effective} information.

In appendix \ref{AppB}, we present further analysis based on the average confidence scores given to individual pieces of model evidence (rather than the entire sequences).

\section{Conclusion}

In this work we take a first step toward addressing a fundamental limitation of existing ML interpretability research. While important progress has been made towards developing interpretability modules to increase the comprehensibility of black-box ML models, these modules have not been empirically tested by end-users who must trust the underlying models in practice. We have proposed an approach to designing human-interpretable systems using RL to \emph{learn what is interpretable} to users.
To demonstrate our approach, we designed a ML-driven DSS providing medical risk assessment, and collected feedback about the system from both clinicians and ML experts. 

Our results provide important insights into interpretability, revealing that system designers cannot predict what information will build the end-user's trust in the system, creating a significant barrier to ML models being used in practice.
Our next step is to recruit more clinicians and ML experts to interact with our DSS so we can obtain statistically significant results.

There are many opportunities for future research. Our proposed framework could be applied for different ML models (e.g., random forests or SVMs), different interpretability modules (e.g., associated classifiers \cite{luo_2016} or DeepLIFT \cite{DBLP:journals/corr/ShrikumarGK17}), 
different applications (e.g., finance or autonomous vehicles), or even different user interfaces, all of which we believe may affect interpretability. We foresee this work as but a first step in the right direction -- learning what is interpretable using sophisticated RL techniques to design trustworthy ML-driven DSSs that clinicians are willing to use in practice. 

\newpage

\bibliographystyle{ieeetr}
\bibliography{MLinDSSBib}{}

\begin{thebibliography}{10}

\bibitem{waljee_higgins_2010}
A.~K. Waljee and P.~D.~R. Higgins, ``Machine learning in medicine: A primer for
  physicians,'' {\em The American Journal of Gastroenterology}, vol.~105,
  no.~6, p.~1224–1226, 2010.

\bibitem{esteva_2017}
A.~Esteva, B.~Kuprel, R.~A. Novoa, J.~Ko, S.~M. Swetter, H.~M. Blau, and
  S.~Thrun, ``Dermatologist-level classification of skin cancer with deep
  neural networks,'' {\em Nature}, vol.~542, no.~7639, p.~115–118, 2017.

\bibitem{gulshan_2016}
V.~Gulshan, L.~Peng, M.~Coram, M.~C. Stumpe, D.~Wu, A.~Narayanaswamy,
  S.~Venugopalan, K.~Widner, T.~Madams, J.~Cuadros, and et~al., ``Development
  and validation of a deep learning algorithm for detection of diabetic
  retinopathy in retinal fundus photographs,'' {\em Jama}, vol.~316, no.~22,
  p.~2402, 2016.

\bibitem{cleophas_zwinderman_2015}
T.~J. Cleophas and A.~H. Zwinderman, {\em Machine Learning in Medicine - a
  Complete Overview}.
\newblock Springer International Publishing Switzerland, 2015.

\bibitem{kononenko_2001}
I.~Kononenko, ``Machine learning for medical diagnosis: history, state of the
  art and perspective,'' {\em Artificial Intelligence in Medicine}, vol.~23,
  no.~1, p.~89–109, 2001.

\bibitem{cabitza_rasoini_gensini_2017}
F.~Cabitza, R.~Rasoini, and G.~F. Gensini, ``Unintended consequences of machine
  learning in medicine,'' {\em Jama}, vol.~318, p.~517, Aug 2017.

\bibitem{obermeyer_emanuel_2016}
Z.~Obermeyer and E.~J. Emanuel, ``Predicting the future — big data, machine
  learning, and clinical medicine,'' {\em New England Journal of Medicine},
  vol.~375, no.~13, p.~1216–1219, 2016.

\bibitem{DBLP:journals/corr/abs-1802-01933}
R.~Guidotti, A.~Monreale, S.~Ruggieri, F.~Turini, D.~Pedreschi, and
  F.~Giannotti, ``A survey of methods for explaining black box models,'' {\em
  CoRR}, vol.~abs/1802.01933, 2018.

\bibitem{ribeiro_singh_guestrin_2016}
M.~T. Ribeiro, S.~Singh, and C.~Guestrin, ``"why should i trust you?"
  explaining the predictions of any classifier,'' {\em Proceedings of the 22nd
  ACM SIGKDD International Conference on Knowledge Discovery and Data Mining -
  KDD 16}, 2016.

\bibitem{slim}
W.~R. Zame, J.~Yoon, F.~W. Asselbergs, and M.~van~der Schaar, ``Interpretable
  machine learning identifies risk predictors in patients with heart failure,''
  {\em American Heart Association (AHA) Scientific Sessions}, 2018.

\bibitem{Craven:1995:ETR:2998828.2998832}
M.~W. Craven and J.~W. Shavlik, ``Extracting tree-structured representations of
  trained networks,'' in {\em Proceedings of the 8th International Conference
  on Neural Information Processing Systems}, NIPS'95, (Cambridge, MA, USA),
  pp.~24--30, MIT Press, 1995.

\bibitem{luo_2016}
G.~Luo, ``Automatically explaining machine learning prediction results: a
  demonstration on type 2 diabetes risk prediction,'' {\em Health Information
  Science and Systems}, vol.~4, Aug 2016.

\bibitem{Yin2018}
M.~Yin, J.~W. Vaughan, and H.~Wallach, ``Does stated accuracy affect trust in
  machine learning algorithms?,'' {\em CoRR}, vol.~abs/1806.07004, 2018.

\bibitem{Collaris2018}
D.~Collaris, L.~Vink, and J.~van Wijk, ``Instance-level explanations for fraud
  detection: A case study,'' {\em CoRR}, vol.~abs/1806.07129, 2018.

\bibitem{pocock_etal._2012}
S.~J. Pocock, C.~A. Ariti, J.~J. Mcmurray, A.~Maggioni, L.~Køber, I.~B.
  Squire, K.~Swedberg, J.~Dobson, K.~K. Poppe, G.~A. Whalley, and et~al.,
  ``Predicting survival in heart failure: a risk score based on 39 372 patients
  from 30 studies,'' {\em European Heart Journal}, vol.~34, no.~19,
  p.~1404–1413, 2012.

\bibitem{buuren_groothuis-oudshoorn_2011}
S.~V. Buuren and K.~Groothuis-Oudshoorn, ``Mice: Multivariate imputation by
  chained equations in r,'' {\em Journal of Statistical Software}, vol.~45,
  no.~3, 2011.

\bibitem{best-arm-identification-multi-armed-bandits}
J.-Y. Audibert, S.~Bubeck, and R.~Munos, ``Best arm identification in
  multi-armed bandits,'' January 2010.

\bibitem{auer2002}
P.~Auer, N.~Cesa-Bianchi, and P.~Fischer, ``Finite-time analysis of the
  multiarmed bandit problem,'' {\em Machine Learning}, vol.~47, no.~2,
  p.~235–256, 2002.

\bibitem{master_clinician}
``Machine learning in prognostic decision-support systems study.''
  \url{https://mlindecisionsupport.github.io/}.
\newblock Accessed: 2018-09-26.

\bibitem{master_cs}
``Machine learning in prognostic decision-support systems study - computer
  science perspective.'' \url{https://mlinterpretability.github.io/}.
\newblock Accessed: 2018-09-26.

\bibitem{miller1956magical}
G.~A. Miller, ``The magical number seven, plus or minus two: Some limits on our
  capacity for processing information.,'' {\em Psychological review}, vol.~63,
  no.~2, p.~81, 1956.

\bibitem{DBLP:journals/corr/ShrikumarGK17}
A.~Shrikumar, P.~Greenside, and A.~Kundaje, ``Learning important features
  through propagating activation differences,'' {\em CoRR},
  vol.~abs/1704.02685, 2017.

\bibitem{karim2018}
A.~Karim, A.~Mishra, M.~H. Newton, and A.~Sattar, ``Machine learning
  interpretability: A science rather than a tool,'' {\em CoRR},
  vol.~abs/1412.6980, Jul 2018.

\end{thebibliography}

\appendix

\section{Appendix: Definitions of Interpretability}
\label{AppInt}

Currently, there is no universally accepted definition of interpretability \cite{karim2018}. A recent comprehensive survey of ML interpretability methods defines ML interpretability as the extent to which ``the model and/or its predictions are human understandable,'' and highlights two additional desiderata for interpretable models: accuracy and fidelity \cite{DBLP:journals/corr/abs-1802-01933}.

We believe there is an important factor missing from the discussion, which stems from the very goal of ML modeling: producing \textit{trustworthy} predictions. A ML model that has high fidelity, high accuracy, and provides highly understandable predictions has limited value if its target users, such as clinicians, do not trust it and are thus unwilling to readily use the model and its predictions in practice.

\begin{figure}[!h]
\centering
\includegraphics[width=0.5\columnwidth]{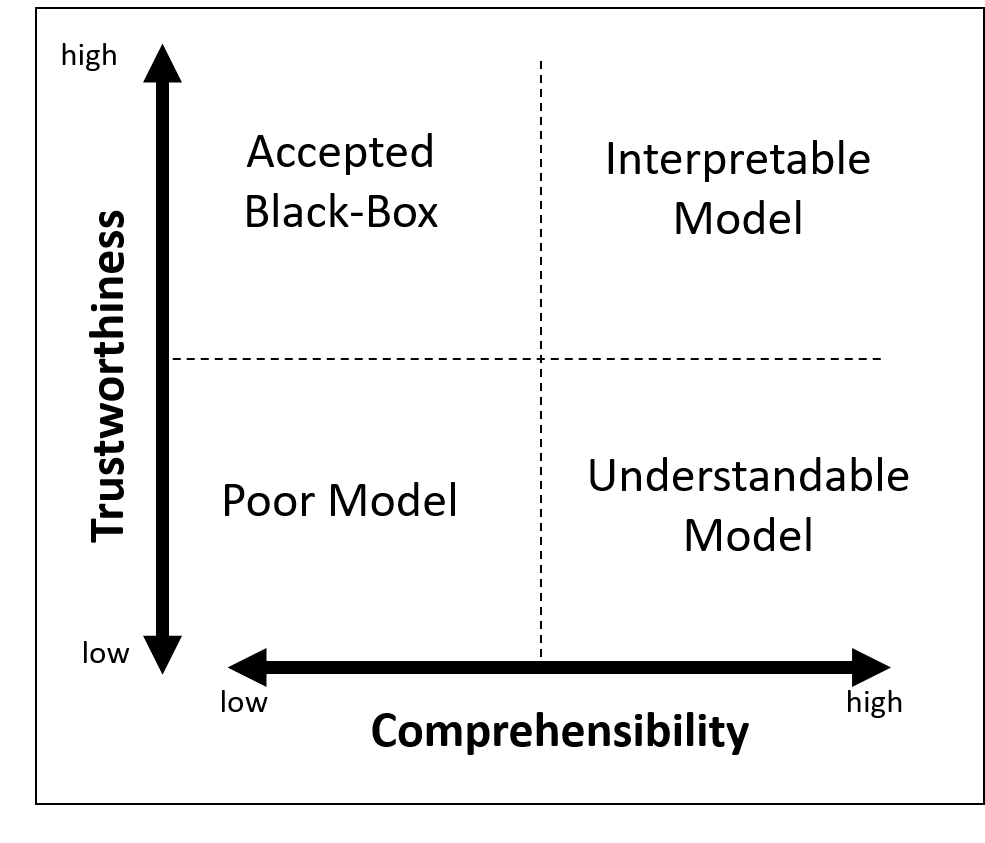}
\caption{The two dimensions of interpretability: comprehensibility and trustworthiness. While current efforts only focus on comprehensibility, we argue that both dimensions are required for interpretability to be achieved.}
\label{fig:IntDim}
\end{figure}

As seen in Fig. \ref{fig:IntDim}, our model of interpretability comprises two dimensions: comprehensibility and trustworthiness. Existing interpretability modules address the first dimension \cite{DBLP:journals/corr/abs-1802-01933}, but neglect to verify that the corresponding ML models will be trusted and utilized by their intended users. %(Note that our interpretability model assumes that the ML model has high accuracy and fidelity.)

Our approach to interpretability addresses this shortcoming. Having developed a deep neural network model (see Section~\ref{DSS}) and corresponding interpretability modules, including local decision-trees and linear model approximations (see Appendix~\ref{AppA}), we explicitly measure user trust/confidence through our experiments. This allows us to attain a model that is both comprehensible and trustworthy, and therefore truly interpretable.

\section{Appendix: Study Participants}
\label{contributers}

%Our study presented an opportunity to test the impact of machine learning interpretability on the confidence of clinicians. As outlined in Section \ref{DSS}, our approach included designing a DSS based on a neural network model predicting risk of mortality for heart failure patients, and using RL to optimize the evidence presented as part of the DSS. The learning was performed by presenting different sequences of evidence to medical experts.

We recruited multiple experts from a variety of fields of expertise, institutions, and nations. Our efforts are a first-step towards an extensive and important collaboration between the ML and medical research communities globally.

Below we present details on some of the collaborators who have either participated as experts commenting on our experimental design and providing \emph{expert heuristics}, or else contributed by completing our online survey tool and providing valuable data-points from which our algorithm could learn the optimal path to interpretability. Additional medical experts have participated anonymously.

%\caption{Partial list of study contributers}

%\label{table:Contributers}
 
%\begin{table}[h]
%\caption{Partial list of study contributers}
%\centering
%\resizebox{1\columnwidth}{!}{
%\begin{tabular}{|l|l|l|l|}
%\hline
%\rowcolor[HTML]{C0C0C0} 
%{\color[HTML]{333333} \textbf{Profile and Expertise}} & {\color[HTML]{333333} \textbf{Affiliation}} & {\color[HTML]{333333} \textbf{Nation}} \\ \hline
%MD, Cardiology & University of Bologna & Italy \\ \hline
%MD, Cardiology, Internal Medicine & University of California, Los Angeles & United States \\ \hline
%MD, Experimental, Diagnostic and Specialty Medicine & University of Bologna & Italy \\ \hline
%MD, Neonatal Medicine & Health Education England - National Health Services & United Kingdom \\ \hline
%MD, Immunology & University of Cambridge & United Kingdom \\ \hline
%MSc, Biomedical and Behavioral Research & San Benedetto Hospital & Italy \\ \hline
%MD, Gastroenterology & University College London & United Kingdom \\ \hline
%MD, Primary Care, Public Health & National Health Services & United Kingdom \\ \hline
%MD, Hospitalist, Internal Medicine & University of California, Los Angeles & United States \\ \hline
%\end{tabular}
%}
%\label{table:Contributers}
%\end{table} 

\begin{table}[h]
	\caption{Partial list of study contributers}
	\centering
	\resizebox{1\columnwidth}{!}{
		\begin{tabular}{|l|l|}
			\hline
			\rowcolor[HTML]{C0C0C0} 
			{\color[HTML]{333333} \textbf{Profile and Expertise}} & {\color[HTML]{333333} \textbf{Nation}} \\ \hline
			MD, Cardiology  & Europe \\ \hline
			MD, Cardiology, Internal Medicine & USA \\ \hline
			MD, Experimental, Diagnostic and Specialty Medicine & Europe \\ \hline
			MD, Neonatal Medicine & Europe \\ \hline
			MD, Immunology & Europe \\ \hline
			MSc, Biomedical and Behavioral Research & Europe \\ \hline
			MD, Gastroenterology & Europe \\ \hline
			MD, Primary Care, Public Health & Europe \\ \hline
			MD, Hospitalist, Internal Medicine & USA \\ \hline
		\end{tabular}
	}
	\label{table:Contributers}
\end{table}

\section{Appendix: Model Evidence Details}
\label{AppA}

The full list of model evidence included in our DSS and experiment is specified in Table \ref{table:InfoSummary}. This evidence was arranged into arms as part of our multi-armed bandit approach to reduce sample complexity. Table \ref{table:Arms} summarizes the various arms tested. Note that the tested arms were not exhaustive -- some arms were removed based on consultations with 3 medical experts, and we did not test the effect of order on the sequence of evidence as this would increase the number of arms exponentially. Future experiments may do well to test this factor.

\begin{table}[h]
\caption[Summary of evidence sequences presented as part of our study]{A detailed summary of the set of evidence sequences
%, or decision-system support design choices $\Omega$, % NHM: I remove this because we don't have this notation anywhere in the paper 
and their associated model information we have included in our study. Evidence is in the order that we present it to users.}
\centering
\resizebox{0.75\columnwidth}{!}{
\begin{tabular}{|l|l|l|l|l|l|l|l|l|}
\hline
{\ul \textbf{Part 1}} & \multicolumn{8}{|l|}{{\textbf{Evidence Sequence}}} \\ \hline
\rowcolor[HTML]{C0C0C0} 
\textbf{Evidence Type} & \textbf{A} & \textbf{B} & \textbf{C} & \textbf{D} & \textbf{E} & \textbf{F} & \textbf{G} & \textbf{H} \\ \hline
\textbf{Data} & X & X & X & X & X & X & X & X \\ \hline
\textbf{Model Training and Implementation} &  &  &  &  & X & X & X & X \\ \hline
\textbf{Model Accuracy} & X & X & X & X & X & X & X & X \\ \hline
\textbf{Linear Approximations Coefficients} &  &  & X & X &  &  & X & X \\ \hline
\textbf{Decision-Tree Approximations} &  & X &  & X &  & X &  & X \\ \hline
\multicolumn{9}{|l|}{{}}  \\ \cline{1-7}
{\ul \textbf{Part 2}} & \multicolumn{6}{|l|}{{\textbf{Evidence Sequence}}} & \multicolumn{2}{|l|}{} \\ \cline{1-7}
\cellcolor[HTML]{C0C0C0}\textbf{Evidence Type} & \cellcolor[HTML]{C0C0C0}\textbf{A} & \cellcolor[HTML]{C0C0C0}\textbf{B} & \cellcolor[HTML]{C0C0C0}\textbf{C} & \cellcolor[HTML]{C0C0C0}\textbf{D} & \cellcolor[HTML]{C0C0C0}\textbf{E} & \cellcolor[HTML]{C0C0C0}\textbf{F} & \multicolumn{2}{l|}{} \\ \cline{1-7}
\textbf{Patient Information} & X & X & X & X & X & X & \multicolumn{2}{l|}{} \\ \cline{1-7}
\textbf{Sensitivity/Interactive Component} &  & X & X & X & X & X & \multicolumn{2}{l|}{} \\ \cline{1-7}
\textbf{Local Linear Model Coefficients} &  &  &  & X &  & X & \multicolumn{2}{l|}{} \\ \cline{1-7}
\textbf{Local Decision-Tree Diagram} &  &  &  &  & X & X & \multicolumn{2}{l|}{} \\ \cline{1-7}
\textbf{Patient Outcome} &  &  & X & X & X & X & \multicolumn{2}{l|}{\multirow{-6}{*}{}} \\ \hline
\end{tabular}
}
\label{table:Arms}
\end{table} 

\begin{figure}[!h]
\centering
\subfloat[Patient scenario]{\includegraphics[height=4cm, valign=b]{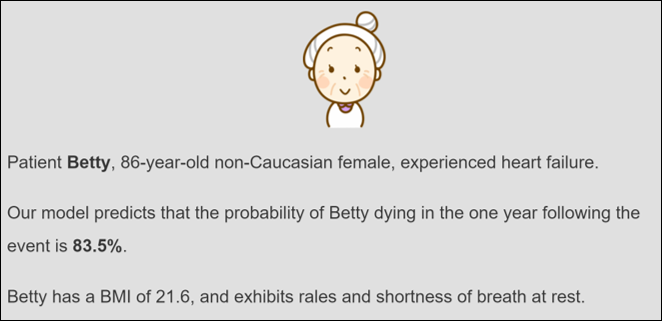}%
	\label{fig:PatientScenario}}
\hfill
\subfloat[Local linear approximation (SLIM)]{\includegraphics[height=4cm, valign=b]{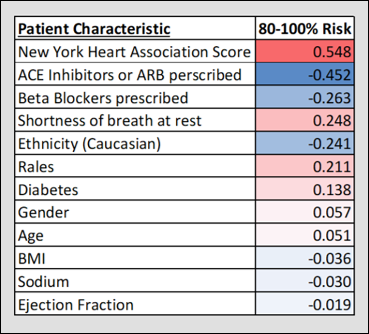}%
	\label{fig:SLIMSample}}
\vfill
\subfloat[Decision-tree approximation]{\includegraphics[height=3.5cm, valign=b]{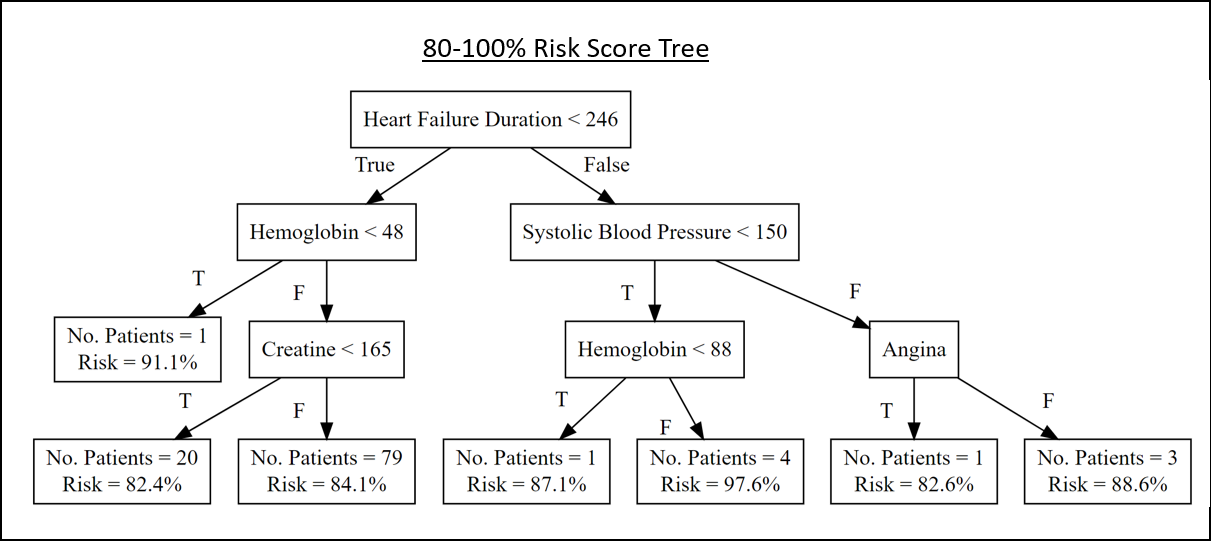}%
	\label{fig:TreeSample}}
\hfill
\subfloat[Feature sensitivity]{\includegraphics[height=3.5cm, valign=b]{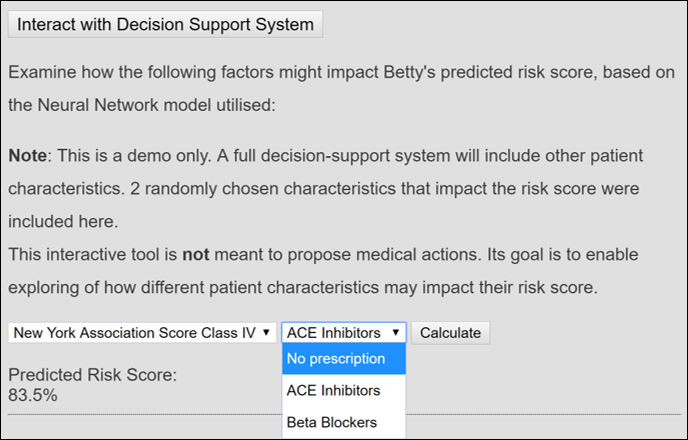}%
%\subfloat[Patient Scenario]{\includegraphics[width=0.5\columnwidth, valign=b]{PatientScenario}%
%\label{fig:PatientScenario}}
%\hfill
%\subfloat[SLIM]{\includegraphics[width=0.25\columnwidth, valign=b]{SLIMSample}%
%\label{fig:SLIMSample}}
%\vfill
%\subfloat[Feature Sensitivity]{\includegraphics[width=0.5\columnwidth, valign=b]{TreeSample}%
%\label{fig:TreeSample}}
%\hfill
%\subfloat[MD: Linear Model]{\includegraphics[width=0.35\columnwidth, valign=b]{FeatureSensitivity}%
\label{fig:Sensitivity}}
\caption{Example patient scenario and interpretability modules presented in our experiments. (a) Sample patient scenario. (b) Local linear approximation module. (c) Local decision-tree approximation module. (d) Interactive feature sensitivity module.}
\label{fig:EvidenceShown}
\end{figure}

Some of the model evidence chosen to display to prospective users in order to increase their confidence in our neural network is presented in Fig. \ref{fig:EvidenceShown}. This evidence includes individual patient scenarios and three interpretability modules, i.e., local linear model approximations, local decision-tree approximations, and a feature sensitivity component. 

We presented four individual patient scenarios to every user. These patients were not part of the training data-set, and are shown with salient characteristics such as age, gender, medical conditions and prescribed medications. They have been embellished with fictional patient names and images, as seen in Fig. \ref{fig:PatientScenario}, to make them more personally relevant to doctors. The patients’ predicted risk-scores range from 15\% to 85\%, and they are representative of the data-set and overall population. Each patient scenario was accompanied with up to three intepretability modules.

Linear model approximations were trained using the Stratified Linear Model (SLIM) methodology \cite{slim}, which improves upon Local Interpretable Model-agnostic Explanations (LIME) \cite{ribeiro_singh_guestrin_2016}. We first used our neural network to partition the training set into risk strata, electing equal quintiles of 0-20\%, 20-40\%, etc. Then, a separate linear regression model was trained for each stratum, and the significant coefficients were presented. Fig. \ref{fig:SLIMSample} represents the linear model coefficients for the 80-100\% stratum. Decision-tree approximations were similarly trained on risk quintiles, and were capped at a depth of 3. Fig. \ref{fig:TreeSample} represents the local decision-tree for the 80-100\% stratum.

%Scenarios may be accompanied by decision-tree or linear model approximation pertaining to the particular data-point's risk stratum, a summary of the patient's true outcome compared with the model's prediction, as well as an interactive sensitivity component.
An interactive feature sensitivity module was implemented as a drop-box, as shown in Figure \ref{fig:Sensitivity}, which allows doctors to interact with the DSS by observing how two randomly selected features can affect the patient's predicted risk. The values were pre-selected and pre-calculated based on the neural network created in Section \ref{DSS}. We did not use the full list of features and possible values due to the computational complexity of re-generating a neural network prediction as part of a web survey, but this is not necessarily a limitation of our approach and future implementations may allow for more sophisticated user interactions with the DSS.

\section{Appendix: Individual Evidence Analysis}
\label{AppB}

\begin{figure}[!h]
\centering
\subfloat[Part 1]{\includegraphics[width=0.47\columnwidth]{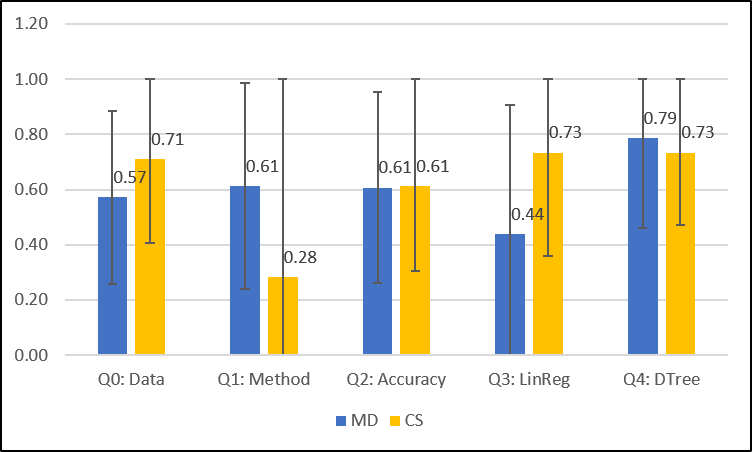}%
\label{fig:RperQ2}}
\hfil
\subfloat[Part 2]{\includegraphics[width=0.47\columnwidth]{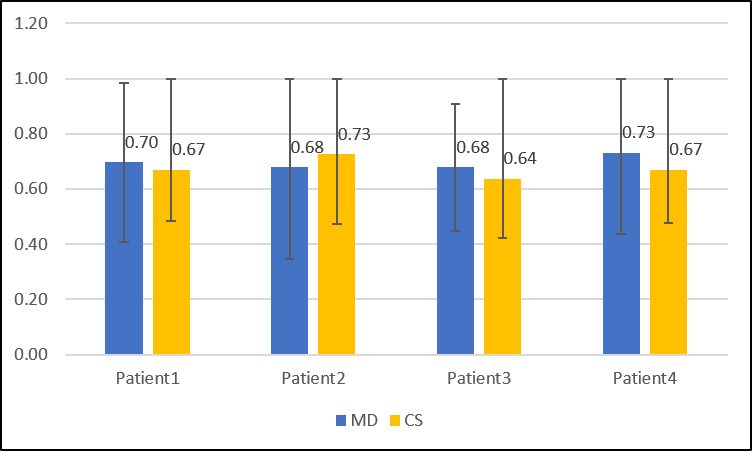}%
\label{fig:RperQ}}
\caption{\ref{fig:RperQ2} shows average rating per individual piece of general model evidence from Part 1, for each user context. The upper and lower bars represent the 95\% confidence interval, capped at 0 and 1. \ref{fig:RperQ} shows average ratings for each patient in Part 2}
\end{figure}

In addition to evaluating the full sequences of model evidence captured by each arm, as shown in Fig. \ref{fig:ResultsArms}, this appendix presents an examination of the effects of each individual type of model evidence on user confidence. In Part 1, we ask respondents to rate how useful they find each piece of model evidence that they encounter. Fig. \ref{fig:RperQ2} summarizes the average and confidence interval of the responses for clinicians and ML experts for Part 1 of our experiment, which focuses on general model evidence.

Doctors appear to find linear model approximations to our neural network the least useful type of evidence. This is in contrast to computer scientists, who believed linear model approximations will have a much higher impact on doctors' confidence, comparable to decision-trees. This suggests that, although the ML research community has been putting significant effort into constructing linear models that represent close approximations to black-box models \cite{ribeiro_singh_guestrin_2016} \cite{slim} these solutions may only provide limited interpretability in practice.

In addition, computer scientists appear to be skeptical about the value of methodology, giving it a low rating. Conversely, doctors rate information about methodology nearly as high as data-set details, suggesting that doctors find information about model training and architecture useful even as the designers feel this information might be too complicated for non-experts, and thus unnecessary.
    
%Finally, it is important to note that decision-trees have the highest rating for both groups, with a rating nearly twice as high as linear models. Though in contrast with the results of Fig. \ref{fig:ResultsArms2}, this suggests that decision-trees are highly effective tools at providing interpretability and raising doctors' confidence. More data is needed to verify the findings, though this may inform future work on ML Interpretability.\newline

Similarly, we can analyze individual responses from Part 2, per patient. Fig. \ref{fig:RperQ} shows these results. It is interesting to note that there are no clear patterns in this figure, and average confidence values are close both across user contexts and across different patients. The fact that there are no significant differences between the results of ML experts and doctors is particularly surprising, as doctors have much more expertise and understanding of patient characteristics.

Additionally, the lack of trend across patients suggests that showing four patients to each user does not necessary lead to an increase in confidence. This suggests that repeated exposure to multiple sample data-points may not be necessary in building model trust.\newline

Though these findings require validation using additional data, we believe that they may have important implications for future developments in ML interpretability using our RL-based approach.

\end{document}